\documentclass[letterpaper, 10 pt, journal]{ieeeconf}  

\IEEEoverridecommandlockouts                              

\usepackage[backend=biber,style=ieee,citestyle=numeric-comp]{biblatex}
\usepackage{hyperref}
\usepackage{amsmath}
\usepackage{amssymb}
\DeclareMathOperator*{\argmin}{argmin}
\usepackage{epsfig}
\usepackage{xcolor}
\usepackage{graphicx}
\usepackage{subcaption}

\title{\LARGE \bf
SD-DefSLAM: Semi-Direct Monocular SLAM for Deformable and Intracorporeal Scenes
}

\author{Juan J. Gómez Rodríguez*, José Lamarca*, Javier Morlana, Juan D. Tardós and José M.M. Montiel

\thanks{* Both authors contributed equally to this work.}
\thanks{This work was supported by the EU-H2020 grant 863146: ENDOMAPPER, the Spanish government grants PGC2018-096367-B-I00, DPI2017-91104-EXP and the MINECO scholarship BES-2016-078678, and by Aragón government grant DGA\_T45-17R.}
\thanks{The authors are with the Instituto de Investigaci\'on en Ingenier\'ia de Arag\'on (I3A), Universidad de Zaragoza, 
Mar\'ia de Luna 1, 50018 Zaragoza, Spain. E-mail: \{jjgomez,  jlamarca, jmorlana, tardos, josemari\}@unizar.es.} }

\begin{document}

\maketitle
\thispagestyle{empty}
\pagestyle{empty}

\begin{abstract}

Conventional SLAM techniques strongly rely on scene rigidity to solve data association, ignoring dynamic parts of the scene. In this work we present Semi-Direct DefSLAM (SD-DefSLAM), a novel monocular deformable SLAM method able to map highly deforming environments, built on top of DefSLAM \cite{lamarca2019defslam}. To robustly solve data association in challenging deforming scenes, SD-DefSLAM combines direct and indirect methods: an enhanced illumination-invariant Lucas-Kanade tracker for data association, geometric Bundle Adjustment for pose and deformable map estimation, and bag-of-words based on feature descriptors for camera relocation. Dynamic objects are detected and segmented-out using a CNN trained for the specific application domain. 

We thoroughly evaluate our system in two public datasets. The mandala dataset is a SLAM benchmark with increasingly aggressive deformations. The Hamlyn dataset contains intracorporeal sequences that pose serious real-life challenges beyond deformation like weak texture, specular reflections, surgical tools and occlusions. Our results show that SD-DefSLAM outperforms DefSLAM in point tracking, reconstruction accuracy and scale drift thanks to the improvement in all the data association steps, being the first system able to robustly perform SLAM inside the human body.

\end{abstract}

\section{Introduction}
Simultaneous Localization and Mapping (SLAM) and Visual Odometry (VO) are fundamental blocks for many applications like autonomous robots or augmented reality. Existing methods can be classified as indirect or direct depending of the manner they perform data association. On the one hand, indirect methods estimate 3D geometry from a set of matched keypoints along covisible images, minimizing a geometric error. On the other hand, direct methods avoid extracting features, and work directly on pixel intensities to estimate the 3D geometry, optimizing a photometric error. Finally, semi-direct methods extract features and combine both types of errors.

However, regardless of that classification, all methods rely on a simple, yet important assumption: scene rigidity. This assumption greatly simplifies the SLAM and VO problem and perfectly models many of their application domains. Nevertheless, the increasing interest in Minimally Invasive Surgery (MIS) and medical robots has placed in the spotlight the rigidity assumption, as these kinds of applications work on highly deforming scenarios. That is why a new classification arises as rigid and non-rigid methods, the latter assuming that the 3D position of triangulated landmarks can vary over time.

\begin{figure}
    \centering
    \includegraphics[width=\columnwidth]{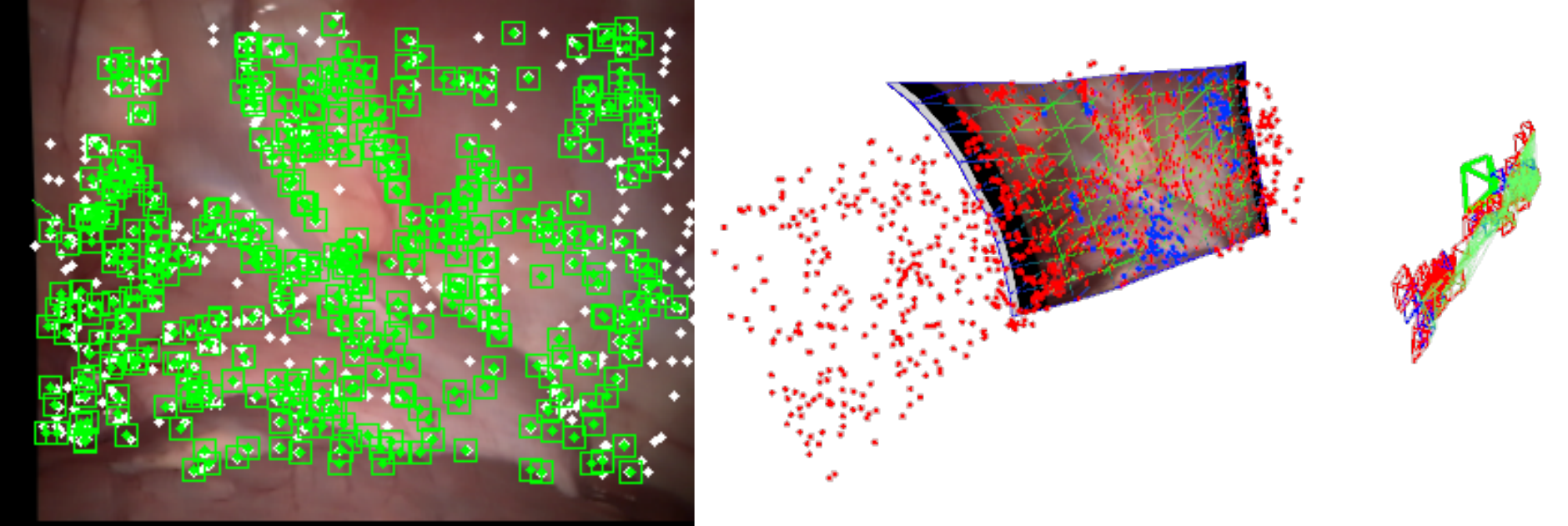}
   \caption{SD-DefSLAM working on Dataset1 of Hamlyn dataset. Left: features tracked in the endoscopic image using photometric techniques. Right: camera motion and growing deformable map estimated by minimizing geometric error.}
    \label{fig:example}
\end{figure}

In this work, building on DefSLAM \cite{lamarca2019defslam}, we propose SD-DefSLAM, the first deformable semi-direct SLAM system, able to robustly process sequences under great deformations and weak texture, as it is the case of MIS videos. SD-DefSLAM is semi-direct as it extracts ORB features and uses an illumination-invariant Lukas-Kanade (LK) \cite{lucas1981iterative} optical flow algorithm to perform data association, minimizing a photometric error, while the camera pose and deforming 3D geometry is estimated minimizing the geometric error (fig. \ref{fig:example}). 

In non-rigid SLAM, dynamic objects are difficult to separate from the deforming background using conventional techniques. To achieve robustness, we mask-out moving objects with the help of a convolutional neural network (CNN) specifically trained to segment surgical tools. Finally, we include relocalization capabilities for which we perform long-term data association with ORB descriptors \cite{rublee2011orb} and a bag of words \cite{GalvezTRO12}, achieving robustness to camera occlusions.


\section{Related Work}
\subsection{Rigid SLAM and VO}
The first real-time SLAM systems followed the indirect approach. MonoSLAM \cite{davison2007monoslam} matches a set of sparse keypoints and recovers the scene geometry in an EFK-based framework. This work was later extended in \cite{civera2008inverse} by using an inverse depth parametrization. Later PTAM \cite{klein2007parallel} proposed a parallelization of the main tasks of an SLAM system to allow a Bundle Adjustment (BA) scheme to optimize the 3D geometry. ORB-SLAM \cite{mur2015orb} is currently the reference system among indirect methods by using the combination of FAST-ORB feature-descriptor \cite{rublee2011orb} and BA to optimize the 3D information. In its successive versions \cite{mur2017orb,campos2020orb} it is extended to different type of sensors, ranging from stereo cameras to wide-lens to inertial sensors. 

As for direct methods, DSO \cite{engel2017direct} is the first fully direct VO algorithm that jointly optimizes structure and motion with photometric BA. This work is later extended in DSM \cite{zubizarreta2020direct} by building a direct SLAM algorithm that uses the same photometric model of DSO. While current direct methods are more robust in weakly textured areas, their accuracy degrades in presence of geometric distortions, and they assume photometric invariance, being only able to adapt to global illumination changes \cite{engel2017direct}. So, they are not applicable in endoscopic images where strong deformations and local illumination changes are prevalent. 

Our work is more similar to SVO \cite{forster2014svo} that proposed an hybrid approach combining direct and indirect methods. SVO is a semi-direct VO method that extracts features in keyframes, uses photometric techniques to perform short-term data association, and ultimately optimizes the reprojection error in a BA. 

The crucial novelty of our method is the use of per-feature illumination-invariant photometric data association, instead of the global image alignment used by DSO and SVO, that cannot handle deforming scenes. Our method also allows to obtain medium-term \cite{campos2020orb} photometric data associations, improving reconstruction accuracy.

\subsection{Deformable SLAM and VO}
Many deformable SLAM and VO systems were developed from rigid ones, aiming in many cases to process intracorporeal sequences, as it is a naturally deforming environment of high practical interest for which several datasets exist \cite{stoyanov2010real,pratt2010dynamic,stoyanov2005soft,Mountney2010ThreeDimensionalTD}. The first systems that processed this kind of images were \cite{grasa2011ekf} and \cite{lin2013simultaneous}, both making use of conventional feature-based SLAM and threshold strategies to differentiate between rigid and non rigid points. Later, ORBSLAM was tuned in \cite{mahmoud2016orbslam} and \cite{mahmoud2017slam} to be able to localize in MIS sequences. 
The seminal work DefSLAM \cite{lamarca2019defslam} is the first indirect monocular SLAM system able to tackle with exploration in deformable scenarios. The system grows the map using a sequential Non-Rigid Structure-from-Motion (NRSfM) algorithm based on \cite{parashar2017isometric}, and estimates at frame rate the deformation occurred and the camera pose by means of a Shape-from-Template (SfT) algorithm \cite{lamarca2018camera}. DefSLAM has been proved to work in some simple medical sequences, but the presence of typical challenges like poor texture, illumination changes and tools intrusion, make it fail. 

This evidences the need of more robust data-association methods to process highly deforming environments. In endoscopic sequences this is usually done by correlation matching in consecutive images \cite{grasa2011ekf}, \cite{lin2013simultaneous}, as feature matching using descriptors such as ORB \cite{rublee2011orb} or SIFT \cite{lowe2004sift} usually do not perform well in low texture regions. AKAZE proposed in \cite{alcantarilla2011fast} is a feature designed to preserve the low texture gradient in the multiscale detector, performing especially well for intracorporeal images. However, it is too slow to be applied in a real-time SLAM algorithm. In \cite{du2015robust}, a deformable Lucas-Kanade \cite{baker2004lucas} implementation is proposed for tracking tissue surfaces in non-exploratory sequences, including a term that controls the deformation. Deep learning techniques can also play an important role as shown in \cite{liu2020extremely} in which they train a CNN to get dense descriptors in a sinus endoscopy dataset.

Finally, as deformable sequences pose a big challenge for SLAM and VO algorithms, it is essential a better understanding of the scene, identifying and removing dynamic objects that could degrade performance. DynaSLAM \cite{bescos2018dynaslam} uses CNNs to detect, remove and inpaint potentialy dynamic objects such as persons or cars. DOT \cite{ballester2020dot} follows up the ideas from DynaSLAM to only mask-out objects that are actually moving. In the case of endoscopic images, the most typical dynamic objects are surgical tool. Segmentation of this kind of objects is of interest to the scientific community and several methods  \cite{laina2017concurrent,kurmann2017simultaneous,pakhomov2019deep} have arisen as response.

\begin{figure}[t]
    \centering
    \includegraphics[width=\columnwidth]{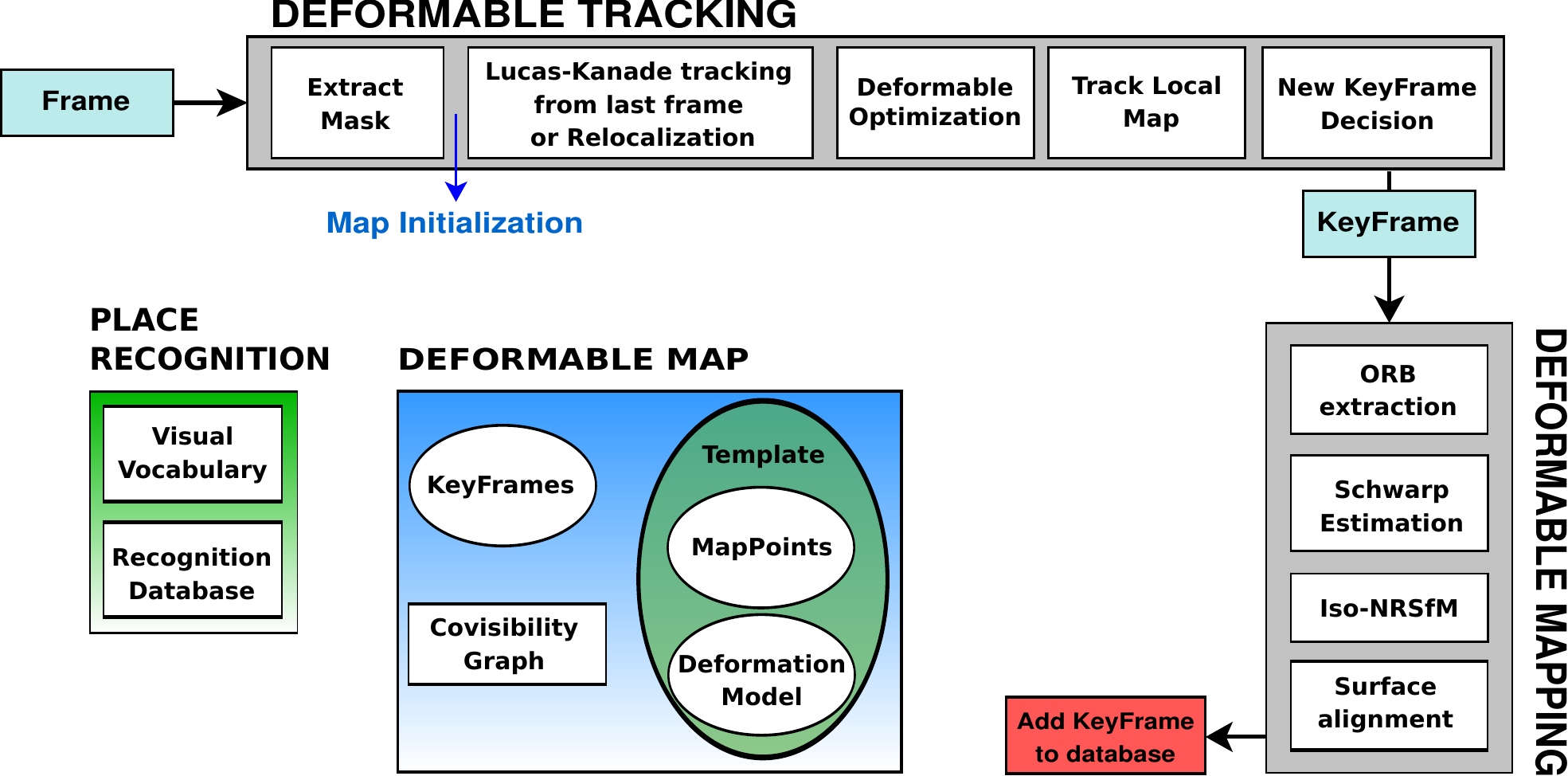}
    \caption{SD-DefSLAM scheme with a tracking and a mapping thread running concurrently. The main novelties are in the tracking thread, that masks surgical tools using a CNN, achieves robustness with an illumination-invariant photometric method that tracks the previous frame and the local map, and includes bag-of-words relocalization and a new regularizer that smooths camera motion.}
    \label{fig:DefSLAM-S_pipeline}
\end{figure}

\section{Semi-Direct DefSLAM}
Our approach is called Semi-Direct Deformable SLAM (SD-DefSLAM) as it performs short-term and medium-term data association \cite{campos2020orb} using a photometric method (subsection \ref{data-association}) while the deformable optimization backend (subsection \ref{def-opt}) optimizes a geometric error. A global overview of SD-DefSLAM is depicted in Fig. \ref{fig:DefSLAM-S_pipeline}. It uses two threads, one for deformable mapping, that progressively builds a growing deformable map and other for deformable tracking, that estimates camera pose and map deformation for each frame processed. Although the main novelties with respect to DefSLAM are in the deformable tracking thread, for the reader convenience, we present here a brief summary of the whole system.


The map is formed by a set of {\em reference keyframes}, that have observed new parts of the scene as exploration progresses, with an associated surface template. Each template models the observed surface with a triangular mesh that represents its shape-at-rest, whose vertices are the 3D map points. The map also contains a set of {\em refining keyframes} that are used to refine the templates. Templates are created and refined by the deformable mapping thread at keyframe rate, and their deformation model is estimated by the deformable tracking thread at frame rate. Keyframes are added to a place recognition database \cite{GalvezTRO12} to enable relocation after occlusions.

The deformation mapping thread estimates the surface observed in the reference keyframes and uses refining keyframes to improve this estimation incrementally. Templates are created to grow the map when exploring new places. The core of the deformation mapping is a Non-Rigid Structure-from-Motion (NRSfM) algorithm  based on isometry and infinitesimal planarity \cite{parashar2017isometric}. It estimates the normal of the points of a keyframe. The points are initialized assuming smoothness in the surface with respect to the rest of normals estimated and they are refined with each new observation. After estimating the normals, a shape-from-normals algorithm estimates a proportional shape of the surface that fits with those normals. Finally, it performs a SE(3) alignment to recover the correct scale with respect to the rest of the map. This new surface becomes the template for the deformation tracking.

The deformation tracking thread estimates the localization of the camera and the deformation of the 3D map surface at frame rate. The map surface is coded by means of its shape-at-rest and a deformation model. The input of the deformation tracking is the last pose of the camera, the last deformation of the template and the new frame. We use a LK tracker to get initial putative matches, that are computed independently for each point. With the putative matches we estimate a initial deformation of the mesh. This optimization is robust to outliers and give us a better estimation of the position of the points. With these new estimates we reinitialize the LK tracker and search for map points in the observed zone. This allow matches with larger baselines than with a standard LK tracker. In case of tracking lost, we have design a relocalization module (subsection \ref{reloc}) able to relocate the system in this map. For our final application, we have incorporated a CNN that segments tools (subsection \ref{dynamic}) to remove matches in dynamic non-modeled objects.


\subsection{Data Association}\label{data-association}

For data association, indirect methods rely on good texture to obtain distinctive features, a RANSAC step to enforce rigidity of the set of matchings found, and robust costs functions in BA to reduce the impact of the remaining outliers. In contrast, direct methods use global image alignment that can use pixels with lower texture but rely even more strongly in scene rigidity. In this section we present a photometric data association method that works reliably in low-textured areas, without relying neither in illumination constancy, not in scene rigidity. For this, we use an enhanced Lucas-Kanade (LK) algorithm to perform short-term data association among all the images in the sequence. Our LK algorithm allows us to track low textured surfaces with subpixel accuracy even though there have been local changes in lighting. Next, we describe the basic LK algorithm to better explain the improvements performed to increase accuracy and robustness. 

\subsubsection{Basic Lucas-Kanade algorithm}
Let be $I$ and $J$ the reference and the current grayscaled images respectively, $\mathbf{u} = (x,y)^T$ a generic image point found in $I$ and $P(\mathbf{u})$ a squared patch centered on $\mathbf{u}$ of size $(2\omega_x + 1) \times (2\omega_y + 1)$ pixels. The goal of LK algorithm is to find the optical flow vector $\mathbf{d} = (d_x,d_y)^t$ such us $I(P({\bf u}))$ and $J(P(\mathbf{u} + \mathbf{d}))$ are similar. This is solved using Gauss-Newton gradient descent non-linear optimization:
\begin{equation}\label{eq::klt_residual}
\argmin_{\mathbf{d}}  \sum_{\mathbf{x} \in P(\mathbf{u})}\left(I(\mathbf{x}) - J(\mathbf{x} + \mathbf{d})\right)^2 
\end{equation}

Note that the goal function depends directly on the gray values of both images and the size of the patch $\omega_x$, $\omega_y$. 

\subsubsection{Enhanced Lucas-Kanade algorithm}
The basic LK optimization (Eq. \ref{eq::klt_residual}) depends directly on the raw intensity values of $I$ and $J$, which makes the LK algorithm very sensitive to illumination changes. While some direct methods address this issue with a global illumination compensation \cite{engel2017direct}, we solve it in a more flexible way using local illumination compensation. In other words, we compute a gain factor $\alpha$ and a bias value $\beta$ per each tracked patch, which are added in the optimization:

\begin{equation}\label{eq::klt_ilu}
\argmin_{\mathbf{d},\alpha,\beta} \sum_{\mathbf{x} \in P(\mathbf{u})}(I(\mathbf{x}) - \alpha J\left(\mathbf{x} + \mathbf{d}) - \beta\right)^2 
\end{equation}

This is especially important when light changes do not occur uniformly across the image, as it happens in outdoor scenes in a cloud-and-clear day, in autonomous car sequences taken during the night, or crucially in endoscopic sequences where the light sources are attached to the endoscope, brightening the image in the areas that get approached, while other areas get darkened. In this cases, global illumination compensation would produce very poor results.

It is also important to keep in mind that the LK algorithm needs the initial guess for $\mathbf{d}$ to be close to the solution in order to converge. That means that if the point to be tracked suffers a big displacement in pixels (for example, due to camera motion or strong deformations) between images, LK may display poor convergence. This can be solved by taking the pyramidal approach proposed in \cite{bouguet2001pyramidal} in which the algorithm estimates the optical flow along a pyramidal representation of $I$ and $J$ from the coarsest to the finest level.

Moreover, as we have geometrical estimates of the 3D scene surface and camera poses provided by the SLAM, we can compute an initial guess for $\mathbf{d}$ using that information to improve convergence. For that purpose, assuming local planarity around each tracked point, we can further compute a homography ($\mathbf{h}$) per point that synthesizes the shape of the patch in the new image, yielding the following error term:

\begin{equation}\label{eq::klt_ilu_h}
\argmin_{\mathbf{d},\alpha,\beta} \sum_{\mathbf{x} \in P(\mathbf{u})}(I(\mathbf{x}) - \alpha J\left(\mathbf{h}(\mathbf{x}) + \mathbf{d}) - \beta\right)^2 
\end{equation}

Transformation defined by $\mathbf{h}$ compensates any rotation or scale change that the patch could have suffered, making our enhanced LK algorithm rotation and scale invariant. It is also essential to note that computations to synthesize the patch use bilinear interpolation to achieve subpixel accuracy. Now the algorithm guesses for $\mathbf{d}$, can be safely set to 0 because most of the flow is estimated from the available geometry. 

Finally, even though LK algorithm converges, it is not guaranteed that it has converged to the correct solution. This can produce spurious feature tracks that negatively affect the overall robustness and accuracy of the algorithm. Most systems address this issue imposing scene rigidity, either in a RANSAC step or with global image alignment. In our case we detect and discard most outliers computing the \textit{Structural Similarity Index} (SSIM) \cite{wang2004image} between the reference and the tracked patches. The remaining outliers are successfully handled by a robust influence function in the deformable optimization.

\subsection{Deformable optimization} 
\label{def-opt}
\begin{figure}
    \centering
    \includegraphics[width=\columnwidth]{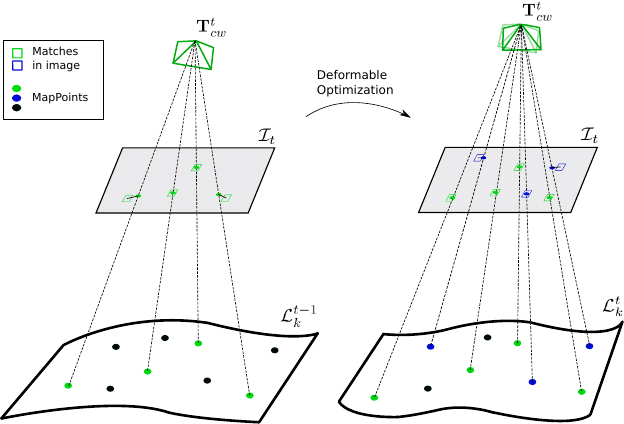}
    \caption{Two-step optimization. First, points from the previous image are tracked using photometric error, getting some matches (green squares) associated to map points (green dots), and we perform a deformable optimization that improves both the camera pose $\mathbf{T}_{cw}^{t}$ and the local map $\mathcal{L}^{t-1}_k$, reducing the geometric error. Then, we project points from the local map $\mathcal{L}^{t}_k$ into the current image, track them them with our enhanced LK method and perform a final deformable optimization. The reobserved map points are marked in blue.}
    \label{fig:two_step}
\end{figure}

Despite our LK algorithm is able to track low textured surfaces in the presence of deformation using photometric error, the innovation between the reprojected map points and their position in the image would be so high that they will be considered as outliers in a pure camera pose optimization. Instead, the tracking thread estimates simultaneously the camera pose and the surface deformation minimizing the geometric reprojection error. This dualism leads to the semi-direct name of our algorithm.

More precisely, our deformable tracking thread performs a two-step optimization (Fig. \ref{fig:two_step}) designed to increase SLAM accuracy by reusing the map. For that purpose, as the camera performs exploration, we compute a local map around the current camera pose with covisible keyframes.

The first step aims to compute a first coarse estimation $\mathbf{T}_{cw}^{t}$ for the camera pose. It obtains putative matches for the pointsin the previous image using the LK tracker with no geometric information (eq. \ref{eq::klt_ilu}) and runs a first deformable pose optimization. With this early optimization, we also compute the local map $\mathcal{M}$ for the next step.

With the computed camera pose $\mathbf{T}_{cw}^{t}$ and local map $\mathcal{M}$ from the previous step, we reproject map points from the local map into the current image. Using the projections and the geometrical information from $\mathcal{M}$, we compute an homography $\mathbf{h}$ per projected point and we search its true image position by running our LK tracker with homographies (eq. \ref{eq::klt_ilu_h}). Finally, with the additional matches found, we run a second deformable pose optimization.

Both deformable optimizations estimate the local map $\mathcal{L}_{k}^{t}$ deformation at frame $t$, along with the camera pose $\mathbf{T}_{cw}^{t}$, using a modified version of the cost function proposed in \cite{lamarca2019defslam}:

\begin{equation}\label{eq:cost_camera}
\begin{split}
 \underset{\mathcal{L}_{k}^t,\mathbf{T}_{cw}^t}{\arg\min} \; \varphi_{d}(\mathcal{I}^t,\mathbf{T}_{cw}^t,\mathcal{L}_{k}^t) & +  \varphi_{e}(\mathcal{L}_{k}^t,\mathcal{L}^{t-1}_k,\mathcal{T}_k) \\ & + \varphi_{c}(\mathbf{T}_{t,t-1})
\end{split}
\end{equation}

\noindent where $\varphi_{d}(\mathcal{I}^t,\mathbf{T}_{cw}^t,\mathcal{L}_{k}^t)$ is the total squared reprojection error weighted with a robust Huber influence function,  and  $\varphi_{e}(\mathcal{L}_{k}^t,\mathcal{L}^{t-1}_k,\mathcal{T}_k)$ is the deformation energy of the template $\mathcal{T}_{k}^{t}$ that considers bending and stretching (see \cite{lamarca2019defslam} for more details). 

To smooth camera motion in frames with low number of matches due to occlusions or sudden deformations, we add here a new regularization term:

\begin{equation}
\label{eq:reg_cam}
\varphi_{c}(\mathbf{T}_{t,t-1}) = 
\xi ^{T} \mathbf{W}  \xi
\end{equation}

\noindent where $\xi = \mathfrak{log} \left(\mathbf{T}_{t,t-1}\right)$ encodes the translation and rotation between the current and previous frame in the Lie algebra, and $\mathbf{W}$ is a hand-tuned information matrix that controls the degree of smoothing performed.

\subsection{Relocalization}\label{reloc}
The presence of deformations, really low textured areas, or complete occlusions can lead to system failure. In that context, it is of paramount importance to have a procedure that allows tracking recovery. As in ORB-SLAM, the detection of candidate keyframes for relocation uses the bag-of-words (BoW) technique from \cite{GalvezTRO12}, building a database with every keyframe in the sequence, converting them into BoW after extracting ORB descriptors. When the system gets lost, we convert the lost frame into BoW and query the recognition database, obtaining some keyframe candidates. For each keyframe, correspondences associated to map points are computed and then, we obtain an initial camera pose with PnP, performing RANSAC iterations. The main difference with the rigid case is that the inlier threshold has been increased to allow points with some deformation. If PnP is successful, we retrieve the template associated with the candidate keyframe and perform a deformable optimization, optimizing both the template and the camera pose. Tracking continues with this retrieved template. Although our method only works under mild deformations, as PnP is constrained by (weakened) rigidity, it is able to successfully solve the typical short-time occlusions appearing in endoscopies.

\subsection{Moving Objects}\label{dynamic}

\begin{figure}[t]
     \centering
     \begin{subfigure}[b]{0.48\columnwidth}
         \centering
         \includegraphics[width=\columnwidth,height=3.5cm,trim={0cm 20px 0cm 0},clip]{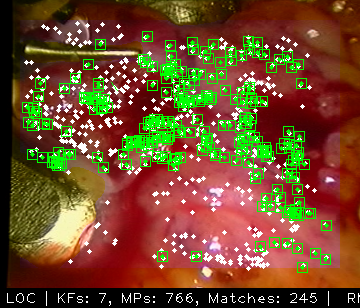}
     \end{subfigure}
     \hfill
     \begin{subfigure}[b]{0.48\columnwidth}
         \centering
         \includegraphics[width=\columnwidth,height=3.5cm,trim={6cm 20px 4.5cm 0},clip]{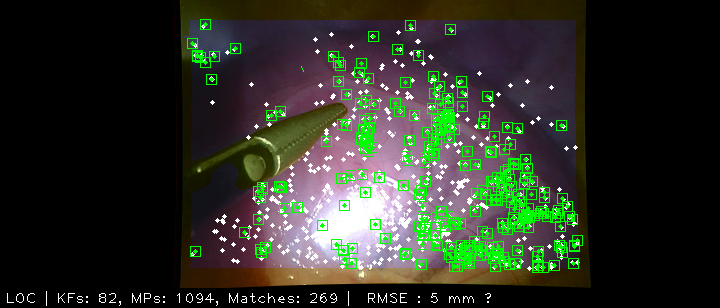}
     \end{subfigure}
     
        \caption{Frames from Hamlin datasets 4 and 19 showing surgical tools, that are successfully detected and masked-out (yellow color) using semantic segmentation with a CNN.}
        \label{fig:masks}
\end{figure}

In conventional SLAM, moving objects can be successfully detected as their motion is not consistent with the motion of the rest of the scene, except if they move too slowly. However, in a deformable scenario, separating object motion from scene deformation is far from trivial using just geometric information. Matches coming from moving objects lead to severe errors in scene deformation or even to total SLAM failure. We propose to solve this issue using semantic information with a CNN trained to identify and segment the typical moving objects in each application domain, masking the corresponding image regions to avoid matching features in them. 

To segment surgical tools in medical scenes we use the CNN defined and trained in \cite{shvets2018automatic}. The network is directly integrated in the system and computes a mask for each incoming image. The mask is finally dilated to avoid keypoint detection in the borders of tools. In Fig. \ref{fig:masks}, we show examples of the masks obtained in two different sequences.

If the tool occludes large parts of the image, the camera pose estimation will become an ill-conditioned problem. For this reason, we constraint the camera motion with a smooth motion prior. When the occlusion is complete, tracking is lost and the system relies on relocation.





\section{Experiments}
We have evaluated the proposed system and compared it with DefSLAM \cite{lamarca2019defslam} in two datasets. The first one is the Mandala dataset created to evaluate deformable SLAM. The purpose of this dataset is to evaluate the performance of the system in a controlled environment with good texture and illumination conditions. Secondly, we further validated our system in several medical sequences of the Hamlyn dataset which pose a substantial challenge to SLAM algorithms. Although our method is pure monocular, in both cases, we use datasets obtained with stereo cameras, to extract a ground truth solution for the scene surface. We analyze the 3D RMS error of the reconstruction, by means of the Euclidean distance between the ground truth and the reconstruction of the system correcting the scale by frame, and reporting the scale drift observed along the trajectory. We also provide a data association quality to compare the performance of the feature matching technique in DefSLAM with the new semi-direct technique that uses photometric information and gives subpixel accuracy.

\subsection{Mandala dataset}

\begin{table}[t]
\centering
\caption{Comparison in Mandala Dataset.} \label{tab:mandala}
\begin{tabular}{c|c|c|c|c|}
\cline{2-5}
& \multicolumn{2}{c|}{DefSLAM \cite{lamarca2019defslam}} & \multicolumn{2}{c|}{SD-DefSLAM}        \\ \cline{2-5} 
& RMSE (mm)      & Scale drift & RMSE (mm)     & Scale drift   \\ \hline
\multicolumn{1}{|c|}{Mandala0} & 26.3 & 1.06 & \textbf{23.1} & \textbf{1.03} \\ \hline
\multicolumn{1}{|c|}{Mandala1} & 22.3 & 1.44 & \textbf{21.3}  & \textbf{1.32}  \\ \hline
\multicolumn{1}{|c|}{Mandala2} & 17.9  & 1.46 & \textbf{16.1} & \textbf{1.41} \\ \hline
\multicolumn{1}{|c|}{Mandala3} & 43.7  & 2.07 & \textbf{41.8} & \textbf{1.26}  \\ \hline
\multicolumn{1}{|c|}{Mandala4} & 55.6  & 1.78  & \textbf{48.1} & \textbf{1.27}  \\ \hline
\end{tabular}
\end{table}

The Mandala dataset consists of 5 sequences exploring a mandala kerchief that goes from a totally rigid situation (Mandala0) to a intensively deforming one (Mandala4). The kerchief is hanged and deformed creating waves that go through it. The intensity of the deformation is measured depending on the speed and amplitude of the waves. 

Table \ref{tab:mandala} shows that SD-DefSLAM  outperforms DefSLAM in all Mandala sequences, both in RMS reconstruction error and in scale drift. While in the most rigid sequence (Mandala0) the improvement is marginal, for those sequences with more aggressive deformations (Mandala3 and Mandala4), SD-DefSLAM achieves a significant improvement.


\subsection{Medical scenes}

\newcommand{\doble}[2]{\begin{tabular}{@{}c@{}}#1\\#2\end{tabular}}
\begin{table}
\centering
\caption{Comparison in Hamlyn Dataset}
\begin{tabular}{l|r|r|r|r|}
\cline{2-5}
 & \multicolumn{2}{c|}{DefSLAM}  & \multicolumn{2}{c|}{SD-DefSLAM}                           \\ \cline{2-5} 
& \multicolumn{1}{l|}{ \doble{RMSE}{(mm)} } & \multicolumn{1}{l|}{ Scale drift } & \multicolumn{1}{l|}{\doble{RMSE}{(mm)}} & \multicolumn{1}{l|}{ Scale drift }  \\ \hline
\multicolumn{1}{|l|}{f5} & 5.00 &  1.01   & \textbf{3.00} &   0.99   \\ \hline
\multicolumn{1}{|l|}{f7} & 4.50 &  0.99   & \textbf{4.35} &   0.99  \\ \hline

\multicolumn{1}{|l|}{Seq\textunderscore heart}  & 3.84  &  2.00   & \textbf{1.17}  & \textbf{1.32}  \\ \hline
\multicolumn{1}{|l|}{Seq\textunderscore abdominal} & 23.98  &  0.98   & \textbf{22.2} &   \textbf{1.01}    \\ \hline
\multicolumn{1}{|l|}{Seq\textunderscore organs} & 13.02 &  1.27   & \textbf{6.63} &   \textbf{1.05}   \\ \hline
\multicolumn{1}{|l|}{Seq\textunderscore exploration} &17.02 &  2.60   & \textbf{12.56} &   \textbf{1.36}   \\ \hline
\end{tabular}
\label{tab:MedicalResults}
\end{table}

We have evaluated our system in several laparoscopic scenes of the Hamlyn dataset. This sequences present a huge variety of scenarios, including phantom hearts with CT ground truth (Dataset11-f5 and Dataset12-f7 in Hamlyn \cite{stoyanov2010real,pratt2010dynamic}), a non-exploratory heart sequence with tool intrusions (Dataset4 - Sequence\textunderscore heart in Hamlyn \cite{stoyanov2005soft}) and three exploratory sequences (Dataset1 - Sequence\textunderscore abdominal, Dataset19 - Sequence\textunderscore organs and Dataset20 - Sequence\textunderscore exploration in Hamlyn \cite{Mountney2010ThreeDimensionalTD}).

\begin{figure*}[t]
    \centering
    \includegraphics[trim = {0cm 0cm 0cm 0cm},width = \textwidth]{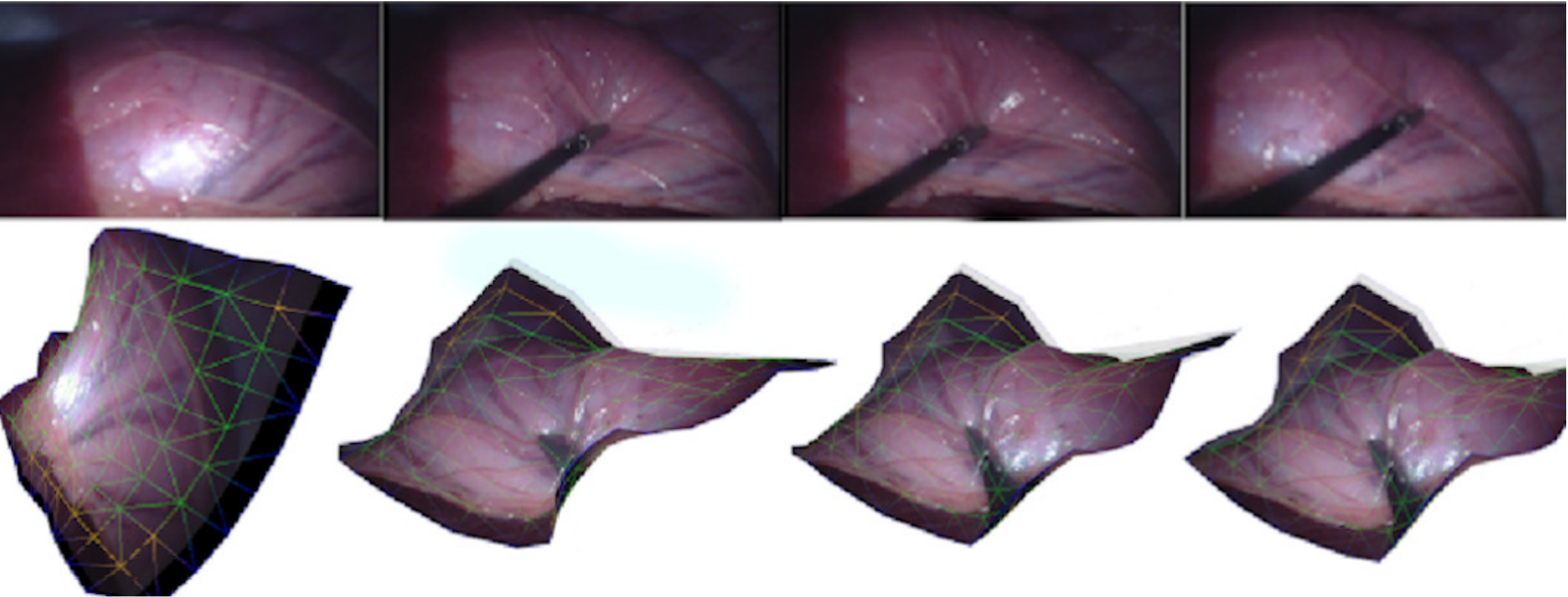}
    \caption{Examples of the reconstructed surfaces in the \textit{Sequence\_organs}. Note how we reconstruct the deformation produced by medical tools. Top is the frame inserted, bottom the 3D reconstruction. From right to left: Frames \#315, \#1010,\#1030,\#1055}
    \label{fig:surfaces}
\end{figure*}

In general, SD-DefSLAM achieves better RMSE and Scale Drift that DefSLAM in all sequences, as shown in Table \ref{tab:MedicalResults}. The improved data association enables our system to better compute the map deformation, improving the RMSE and Scale Drift while the addition of a CNN to mask out surgical tools in the Sequence\textunderscore heart and Sequence\textunderscore organs allows our system to robustly process the sequences with significant improvement in the performance. An example of the reconstructed surfaces under deformations is shown in Fig. \ref{fig:surfaces}.



\subsection{Data association}
The results in the last sections show how SD-DefSLAM outperforms DefSLAM in all the tested datasets. One of the keys for the improvement is the data association which is a fundamental part for both the tracking and the mapping. For the tracking, better association leads to better estimation of the deformations in the map. Concerning the mapping, longer tracks between keyframes speed up the convergence of NRSfM.

In this section, we analyse the proposed data association scheme. There are two key differences \textit{wrt.} the original method. The most evident one is that the matching is performed photometrically, reaching subpixel accuracy. The other one is that each patch is initialized with the keyframes and actively tracked in the consecutive images, removing feature extraction from the matching stage. This is significant as FAST features have low repeatability between temporarily close images, impairing  SLAM performance. 

\begin{figure}
    \centering
    \includegraphics[trim = {4cm 0cm 3cm 0cm},clip,width=\columnwidth]{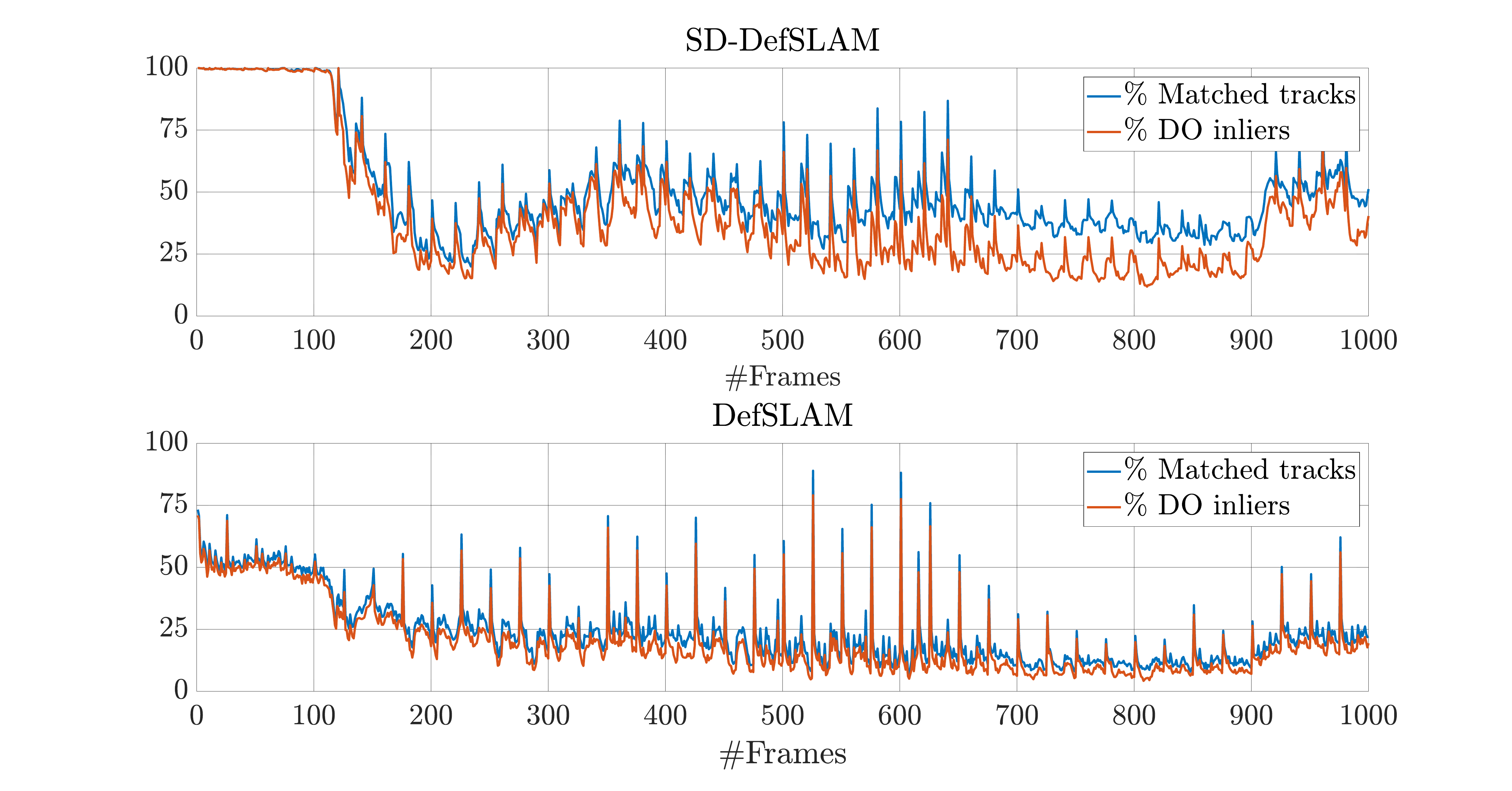}
    \caption{Percentage of points in the local map that are tracked (blue) and that are considered inliers after deformable optimization (orange) in Mandala3 .}
    \label{fig:MatchQuality}
\end{figure}

\begin{figure}
    \centering
    \includegraphics[trim =  {4cm 0cm 3cm 0cm},clip,width=\columnwidth]{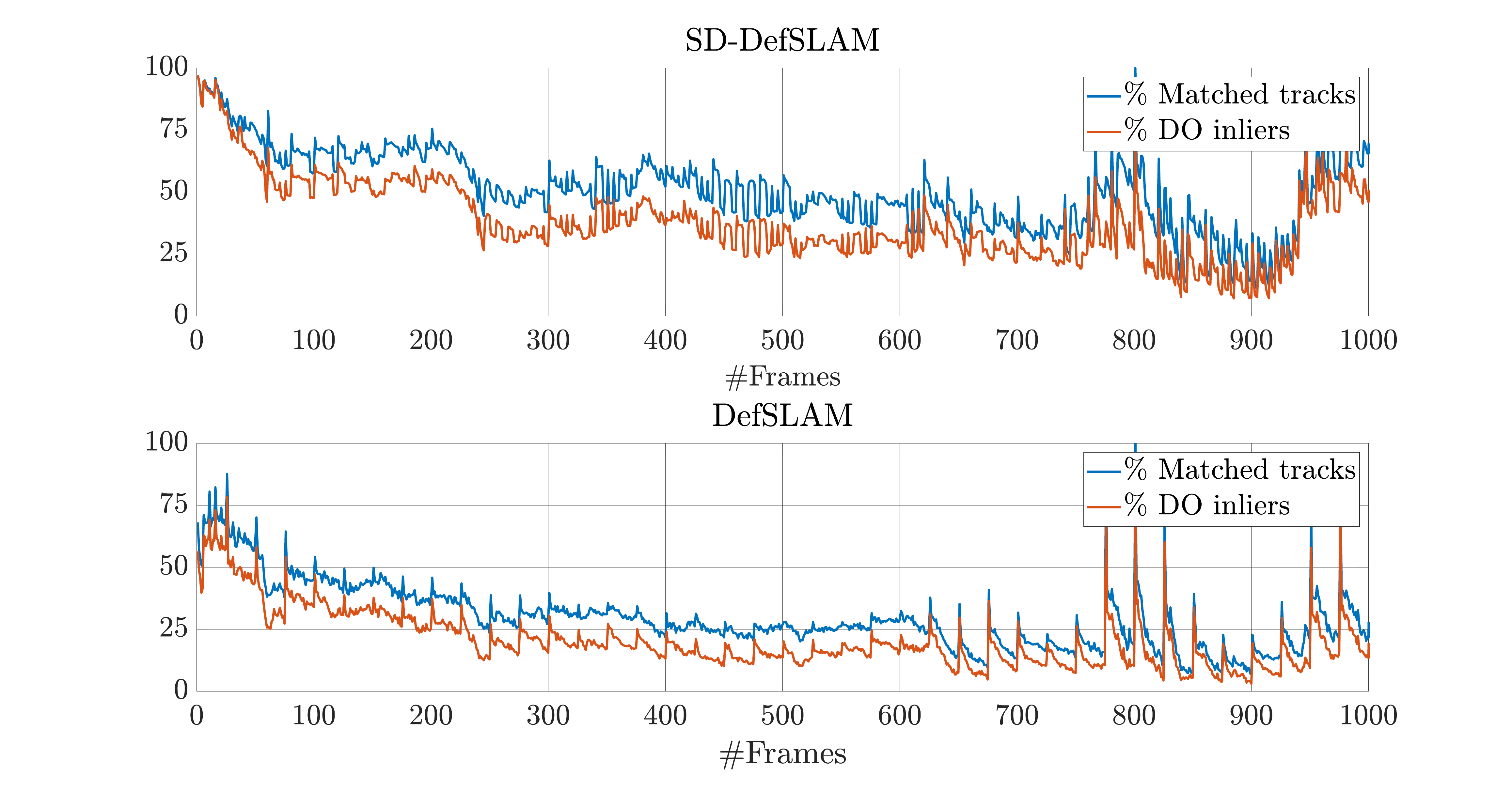}
    \caption{Percentage of points in the local map that are tracked (blue) and that are considered inliers after deformable optimization (orange) in Hamlyn Dataset20.}
    \label{fig:MatchQualityExp}
\end{figure}

Figures \ref{fig:MatchQuality} and \ref{fig:MatchQualityExp} depict a comparison between the SD-DefSLAM photometric data association (top) and the ORB matching of DefSLAM (bottom) in Mandala3 and Hamlyn Dataset20. In both images, the percentage of matched map points (matched tracks) is shown in blue and the inliers after the deformation optimization (DO inliers) in orange. The true positive are the matched map points considered inliers by the deformable optimization, representing the efficiency of the matching system. 

In Mandala3 sequence, SD-DefSLAM doubles the percentage of correct matches obtained by DefSLAM. This greatly improves the overall robustness of the system at the same time that improves the accuracy. Dataset 20 poses a bigger challenge as the combination of low texture and image blurring penalizes both types of data association algorithms, but the new method is still clearly superior. This, together with the subpixel accuracy explains the more accurate reconstruction and smaller scale drift obtained (last row in Table \ref{tab:MedicalResults}).



\subsection{Relocalization}

\begin{figure}
     \centering
     \begin{subfigure}{\columnwidth}
         \centering
         \includegraphics[width=\columnwidth]{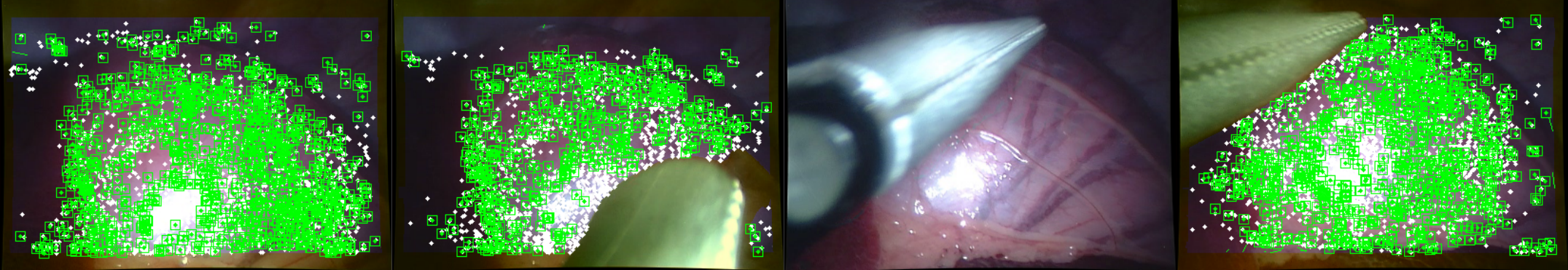}
         \caption{Relocalization in Dataset19.}
         \label{fig:tool_reloc}
     \end{subfigure}
     \newline
     \begin{subfigure}{\columnwidth}
         \centering
         \includegraphics[width=\columnwidth]{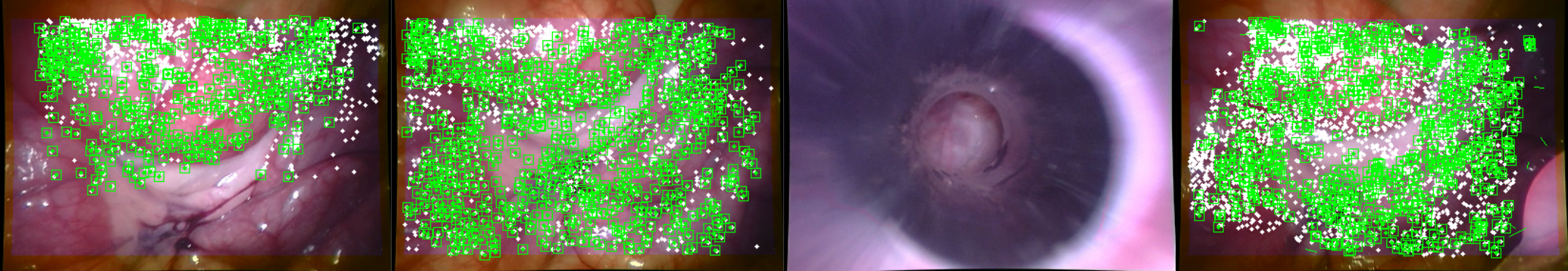}
         \caption{Relocalization in Dataset20}
         \label{fig:outin_reloc}
     \end{subfigure}
     
        \caption{(a) Relocalization due to tool intrusion. The CNN is able to detect the tool correctly, but when it occludes most of the image, the system fails and performs relocalization. (b) In this case, the endoscope is extracted from the scene to clean it, but the system is able to relocate the pose once it is introduced again in the body.}
        \label{fig:reloc}
\end{figure}

Besides the robustness of the system to tools or low texture, the camera still can get totally occluded or even the endoscope must leave the scene to clean the optics. Thanks to the relocalization module, we were able to relocate the system after a tracking failure. In contrast with DefSLAM, which only cannot manage tracking lost, we were able to process more frames in the proposed sequences, see Fig. \ref{fig:reloc}.

\section{Conclusions}

While rigid SLAM is mature, deformable environments pose serious challenges requiring to re-think all data association steps. We have shown that a semi-direct approach based on per-feature  illumination-invariant photometric tracking greatly improves data association, reconstruction accuracy and scale drift. Its combination with CNN segmentation to detect moving objects, and relocalization capabilities to deal with occlusions, gives the first SLAM system able to robustly address the real-life challenges of medical sequences. 



Our deformable model assumes isometric deformations. This is quite a restrictive assumption that is not always fulfilled as is the case of MIS sequences. This causes a worsening in the estimation of the deformation which in turn affects the quality of the data association. This can be addressed by exploring new deformation models that properly represent non-isometric deformations.

\addtolength{\textheight}{-5cm}




\printbibliography

\end{document}